\pdfoutput=1
\documentclass[11pt]{article}

\usepackage{authblk}

\usepackage[]{acl}

\usepackage{times}
\usepackage{latexsym}

\usepackage[T1]{fontenc}
\usepackage[utf8]{inputenc}

\usepackage{microtype}

\usepackage{amsmath}
\usepackage{bbold}
\usepackage{algorithm} 
\usepackage{algpseudocode}
\usepackage{amsmath}
\usepackage{multirow}

\usepackage{booktabs}
\usepackage{url}
\usepackage{graphicx}
\usepackage{subfigure}

\usepackage{color}
\usepackage{soul}
\usepackage[normalem]{ulem}
\usepackage[subtle]{savetrees}
\usepackage{url}
\title{Teaching Neural Module Networks to Do Arithmetic}

\author[ ]{Jiayi Chen}
\author[ ]{Xiao-Yu Guo}
\author[ ]{Yuan-Fang Li}
\author[ ]{Gholamreza Haffari}
\affil[ ]{Faculty of Information Technology, Monash University, Melbourne, Australia}
\affil[ ]{\url{jche0069@student.monash.edu}}
\affil[ ]{\url{{xiaoyu.guo, yuanfang.li, gholamreza.haffari}@monash.edu}}

\begin{document}
\maketitle

\begin{abstract}
Answering complex questions that require multi-step multi-type reasoning over raw text is challenging, especially when conducting numerical reasoning.
Neural Module Networks (NMNs), follow the \textit{programmer-interpreter} framework and design trainable modules to learn different reasoning skills.
However, NMNs only have limited reasoning abilities, and lack numerical reasoning capability.
We upgrade NMNs by: (a) bridging the gap between its \textit{interpreter} and the complex questions; (b) introducing addition and subtraction modules that perform numerical reasoning over numbers.
On a subset of DROP, experimental results show that our proposed methods enhance NMNs' numerical reasoning skills by 17.7\% improvement of F1 score and significantly outperform previous state-of-the-art models.
% Neural Module Networks (NMNs) are an interpretable method for solving Complex Question Answering (CQA) problems. It adopts differentiable neural modules in a programmer-interpreter framework to propagate the probability of each step's symbolic reasoning operation. However, NMNs severely lag in its numerical reasoning capability, a significant aspect of machine reading comprehension. In this paper, we propose two methods for improving NMNs' ability in handling arithmetic operations: (a) incorporate the influence of question information in the interpreter; and (b) introduce multi-numbers addition and subtraction modules to solve more complex arithmetic problems. On a subset of DROP, Experimental results show that our techniques significantly enhance NMNs' performance on arithmetic questions. Our method improves its overall performance by 17.7 absolute points of F1. It also outperforms the strong model NumNet by 2.9 F1 points. 
\end{abstract}

%!TEX ROOT = ./coling2022.tex
\begin{figure*}[hbt!]
  \centering
  \includegraphics[width=1.0 \textwidth]{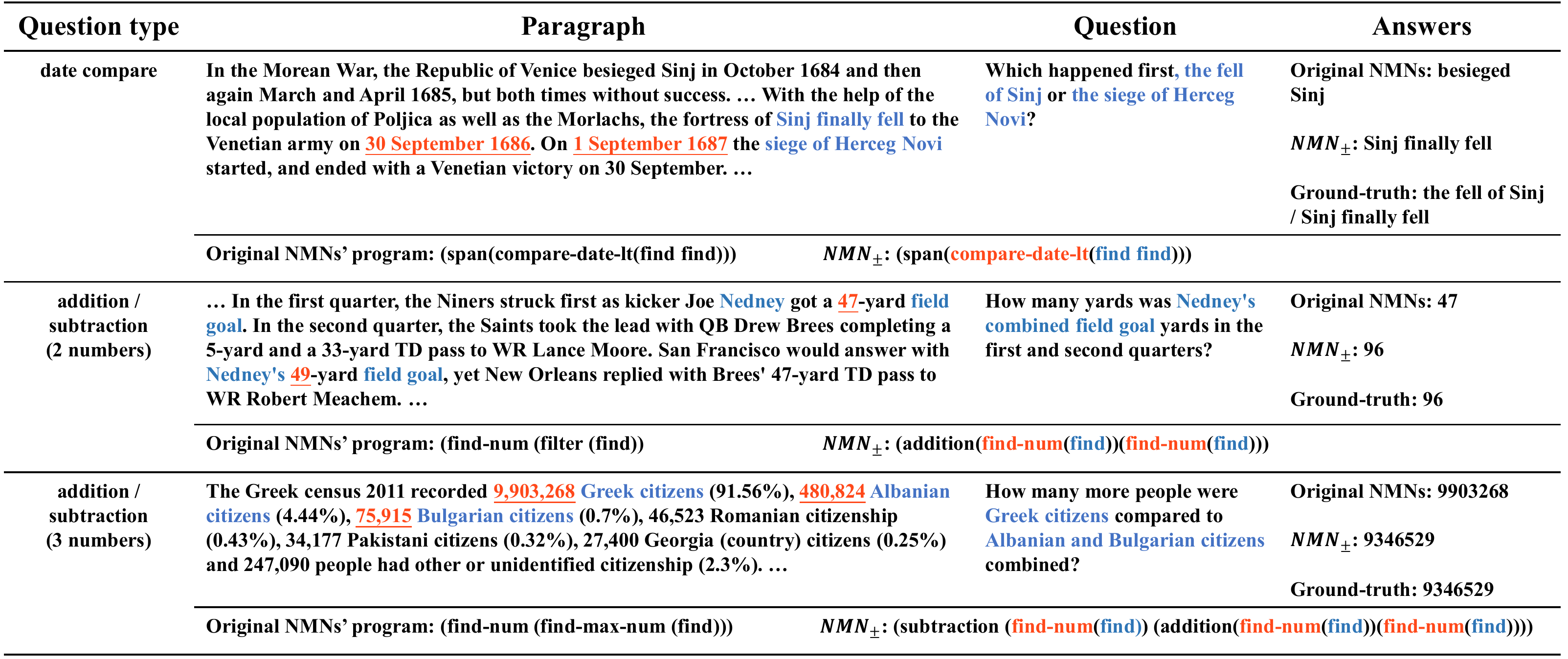}
  \caption{Three examples in the DROP dataset and the predictions by original NMNs and our improved model NMNs$_{\pm}$. The relevant tokens and their corresponding modules are highlighted.}
  \label{fig:main1}
%   \vspace{-8pt}
\end{figure*}

\section{Introduction}
Complex Question Answering (CQA) over text is a challenging task in Natural Language Understanding (NLU). 
% The ``understanding'' processes include but not limit to language embedding, feature extraction, context-question interaction and answer prediction \cite{2019Neural}.
Based on the \textit{programmer-interpreter} paradigm, Neural Module Networks (NMNs) \cite{Gupta2020NeuralMN} learn to first parse complex questions as executable programs composed of various predefined trainable modules, and then execute such programs (implemented by modules) over the given paragraph to predict answers of all kinds. 
NMNs achieve competitive reasoning performance on a subset of DROP \cite{dua2019drop}, and possess remarkable interpretability that is also important for CQA.

However, NMNs' numerical reasoning capability is insufficient: it is incapable of handling arithmetic operations such as addition and subtraction between numbers, which make up nearly 40\% questions of the DROP dataset.
% the scientific, financial and technical text. 
Moreover, a gap exists between the interpreter and the complex question since there is no interaction between them.
% Moreover, other types of numerical reasoning, including comparison and difference between dates, can also be improved, which can be confirmed by our error analysis of the original NMNs on its DROP subset (Appendix \ref{sec:appendix3}). 
Motivated by these, we propose two methods to improve NMNs' numerical reasoning skills. 

First, we incorporate the original question in the interpreter, aiming to \textbf{directly provide question information in the ``execution'' process}, especially number-related questions. 
The intuition behind is that, in the original NMNs, questions participate in the process only through the programmer.
This can cause a distance between queries and returns. 
For example, in Figure \ref{fig:main1}, the first row shows that the original NMNs found the wrong event (i.e., `besieged Sinj') solely based on the paragraph information.
In contrast, our model NMNs$_\pm$ can easily target the correct event (i.e., `Sinj finally fell') with the help of question information.

Second, we introduce new modules to \textbf{support addition and subtraction of up to three numbers}. 
Endowing NMNs with the ability to support arithmetic can greatly boost its overall performance on DROP and beyond. 
For instance, in Figure \ref{fig:main1}, the second row shows that the original NMNs improperly adopt the \verb|find-num| module for the addition question because the module set does not cover such an arithmetic ability. 
To facilitate the learning of the \verb|add/sub| modules, we extract QA pairs related to addition and subtraction from the original DROP dataset to construct a new dataset for training and evaluation.
% and provide heuristically-obtained program supervision for a subset of data (appr. 20\%). 

Experimental results show that our methods significantly enhance NMNs' numerical reasoning capability. 
On a subset of DROP, our methods improve F1 score by 17.7\% absolute points, and on \textsc{add-sub} questions by 65.7\% absolute points. 
Compared to NumNet \cite{2019NumNet}, which is specifically designed for numerical reasoning, our method outperforms it by 2.9\% absolute F1 points.

%!TEX ROOT = ./coling2022.tex
% \vspace{-3pt}
\section{Background and Related Work}
% \textbf{Question answering} (QA) is an important task in NLP. QA models are trained to perform different capabilities to respond to short, precise, and question-specific answers. 

\noindent \textbf{Semantic Parsing} is a widely-adopted approach in the compositional question answering (CQA) task, which involves a number of reasoning steps. In this approach, a \textit{programmer} maps natural-language questions into machine-readable representations (logical forms), which are executed by an \textit{interpreter} to yield the final answer. For instance, WNSMN \cite{2021Weakly} uses a generalized framework of dependency parsing inspired by the Stanford dependency parse tree \cite{chen-manning-2014-fast} to parse queries into noisy heuristic programs. Neural Module Networks \cite{Gupta2020NeuralMN} extend semantic parsing by making interpreter a learnable function with specified modules and executing the logical forms from the programmer in a step-wise manner. % (Figure \ref{fig:main2}).

% \begin{figure}[hbt]
%   \includegraphics[width=\columnwidth]{pdf/Figure_2.pdf}
%   \caption{Semantic parsing of NMNs}
%   \label{fig:main2}
%   \vspace{-12pt}
% \end{figure}

\noindent \textbf{Neural Module Networks} initially is proposed to overcome the Visual Question Answering (VQA) problem \cite{NMN16}, where questions are often compositional. 
\citet{Gupta2020NeuralMN} employs the programmer-interpreter framework with attention \cite{vaswani2017attention} to tackle the CQA task. 
Specifically, the programmer parses each question into an executable program. 
The interpreter takes the program as input and perform various symbolic reasoning functions. 
The modules are defined in a differentiable way, aiming to maintain the uncertainty about each intermediate decision output and propagate them through layers. 
For instance, the predicted program of the first example in Figure \ref{fig:main1} is \verb|span(compare-date-lt(find,find))|. 
The interpreter would first calls the \verb|find| module twice to find events queried by the question (e.g., `the fell of Sinj') and outputs appropriate paragraph attention. 
The \verb|compare-date-lt| module can further locate the dates (e.g., `30 September 1686') to compute their relation. 
By demonstrating the intermediate reasoning steps in this manner, NMNs perform interpretable problem-solving.

\noindent \textbf{Numerical Reasoning} is a necessary ability for models to handle the CQA task \cite{GevaGB20}.
\citet{dua2019drop} modify the output layer of QANet~\cite{YuDLZ00L18-qanet} and propose a number-aware model NAQANet to deal with numerical questions.
NumNet~\cite{2019NumNet} leverage Graph Neural Network to capture relations between numbers.
Similarly, QDGAT~\cite{chen-etal-2020-qdgat} distinguish number types more precisely by adding the connection with entities and obtained better performance.
Nerd~\cite{ChenLYZSL20-nerd} search possible programs exhaustively based on answers and employed these programs as weak supervision.
Another similar work \cite{GuoLH21} proposes a question-aware interpreter but uses a totally different approach to measure the alignment between the question and the context paragraph.
Though these approaches can achieve the high performance on DROP dataset, it is incomprehensible for the reasoning procedure.

%!TEX ROOT = ./coling2022.tex
% \vspace{-4pt}
\section{Model}\label{sec:method}
In this section, we tend to illustrate our proposed methods.
Basically, we will show the incorporation of questions in Section \ref{subsec:question information}.
In Section \ref{subsec:addition and subtraction}, the newly extended module: addition and subtraction will be described.

% \vspace{-2pt}
\subsection{The Incorporation of Questions}
\label{subsec:question information}
Taking one module \verb|compare-date| as a case study: it performs comparisons between two references queried by the question. 
A key reasoning step inside, is the \verb|find-date| module that obtains appropriate a date token distribution $D$ related to each reference: $\verb|find-date|(P) \rightarrow D$.
%This is calculated by the token-to-date similarity matrix $\textbf{S}_{date} = \textbf{P}^{T}_{i:}\textbf{W}_{date}\textbf{P}_{d_j}$, where $\textbf{P}_i$ represents the contextualized embeddings of the $i^{th}$ paragraph tokens and $\textbf{P}_{d_j}$ of the $j^{th}$ date tokens in the paragraph. 
It is worth noting that there is no interaction with the question, which could contain essential information (e.g., entities) that is useful to correctly answer the question.
Therefore, we revise the \verb|find-date| module as follows: $\verb|find-date|(P, Q) \rightarrow D$:
% \vspace{-6pt}
% \begin{align}
%     % C_i            &= [\alpha P_{i}; (1-\alpha )Q_{i}], \label{eq1}\\
%     \textbf{S}^{date}_{i,j} &= [\alpha \textbf{P}_{i}; (1-\alpha )\textbf{Q}_{i}]\textbf{W}_{date}\textbf{P}_{d_j}, \label{eq2}\\
%     \textbf{A}^{date}_{i:}  &= softmax(\textbf{S}^{date}_{i:}), \label{eq3}\\
%     D              &= {\textstyle \sum_{i}} [\alpha P_{i}; (1-\alpha )Q_{i}] \cdot \textbf{A}^{date}_{i:} \label{eq4}
% \end{align}
\begin{align}
    % C_i            &= [\alpha P_{i}; (1-\alpha )Q_{i}], \label{eq1}\\
    \textbf{S}^{date}_{i,d_j} &= [\alpha \textbf{P}; (1-\alpha )\textbf{Q}]_{i}\textbf{W}_{date}\textbf{P}_{d_j}, \label{eq2}\\
    \textbf{A}^{date}_{i:}  &= softmax(\textbf{S}^{date}_{i:}), \label{eq3}\\
    D              &= {\textstyle \sum_{i}} [\alpha P; (1-\alpha )Q]_{i} \cdot \textbf{A}^{date}_{i:} \label{eq4}
\end{align}
where $\textbf{P}$ and $\textbf{Q}$ represent the contextualized embeddings of the paragraph and question, and $\textbf{P}_{d_j}$ of the $j^{th}$ date tokens in the paragraph, $\textbf{W}_{date}$ is a trainable parameter, $P, Q$ are the expected attention distribution of the paragraph and the question respectively. 

In Equation \ref{eq2}, we concatenate the paragraph embeddings $\textbf{P}$ and question embeddings $\textbf{Q}$ that output from a pre-trained BERT \cite{DBLP:conf/naacl/DevlinCLT19} model to construct the context representation.
A hyper-parameter $\alpha$ is used to adjust their contributions, whose value is empirically determined (Appendix \ref{sec:appendix0}). 
The context representation is provided to compute the improved similarity matrix $\textbf{S}_{date}$. 
We concatenate the paragraph and question attention inputs in the same way to calculate the final expected distribution over the date tokens $D$ (Eq.\ \ref{eq4}).
Now the interpreter is equipped with question information to make the prediction. 

\subsection{Addition and Subtraction Modules}
\label{subsec:addition and subtraction}
In the NMNs' modelling paradigm, for addition/subtraction operations, the programmer takes as input two number distributions and produces an output number distribution over all possible result values: $\verb|add/sub|(N_1, N_2) \rightarrow RL$. $N_1$ and $N_2$ represent the probability distributions of the $1^{st}$ and $2^{nd}$ operands over all numbers that are extracted from the paragraph and collected into a sorted operand list $OL$. The positive and negative values of these numbers are exhaustively combined in pairs, from which the possible results of addition/subtraction operations are compiled into a sorted result list $RL$. %The modules take two number distributions as inputs and produce a distribution over the composed-numbers as output. 
For each input number distribution $N_i, i=1,2$, a matrix $\textbf{C}_i\in \mathbb{R}^{m\times n}$ is constructed, where $m$ is the total number of possible results, and $n$ is the maximum number of unique combinations. 
Each value $\textbf{C}_i[j,k]$ is found by looking up the probability value in $N_i[k]$ where $OL[k]$ is the $i^{th}$ operand in any pair that produces result $RL[j]$. The probability that the $j^{th}$ number in $N_i$ is the correct operand of the $k^{th}$ pair. %\sout{The value of $C_i$ is indexed from the probability of corresponding number token distributions.} 
We compute the marginalized joint probability by summing over the product of $C_i$ as the expected distribution over result list $RL$. For the addition module, it is:

\begin{align*}
&p(prediction=RL[j])=\\
&\sum_{k_1,k_2=1}^n \mathbb{1}_{(OL[k_1]+OL[k_2]=RL[j])}\textbf{C}_1[j,k_1]*\textbf{C}_2[j,k_2]
\end{align*}

% \vspace{-8pt}
For instance, assume the sorted operand list $OL$ from a paragraph is [1, 5, 7, 11] and $N_1=[0.1,0.4,0.2,0.3]$. Different combinations are formed, e.g., (+$n_1$, +$n_2$) for addition and (+$n_1$, -$n_2$) for subtraction, and all possible results of the combinations are compiled into two result lists, one for addition and one for subtraction. For subtraction in this case, $RL=[0,2,4,6,10]$. The value of $\textbf{C}_1[2, 1]$ is 0.4, which is found from $N_1[1]$ because the result 4 can be calculated from (+5, -1); and $\textbf{C}_1[2, 3]=0.3$ which equals to $N_1[3]$ as 4 is the result of (+11, -7) as well.
$\textbf{C}_2$ is computed in the same way to further obtain final distribution over $RL$. 

We compose \verb|add/sub| modules in programs to perform 3-number arithmetic. The key to our approach is to construct and distinguish appropriate $C_i$ and $RL$ in different reasoning steps. 
In the second arithmetic step, we should combine the operand list from the paragraph and the result list from the previous step to obtain a new result list $RL'$, $\verb|add/sub|(RL, N) \rightarrow RL'$. Due to the changes in operands and results, the modules should refer to a different $\textbf{C}_i'\in \mathbb{R}^{m'\times n'}$ in the computation.
We extend 2-number \verb|add/sub| modules to recognize the participation of the third number by conditional statement,  in order to differentiate the operand and result lists the interpreter should refer to in different steps.
% Therefore, we define the \hl{new modules} $\verb|add/sub|'(RL, N) \rightarrow RL'$ \yf{how are they different from the existing ones?}, in order to differentiate the arithmetic steps and the operand and result lists the interpreter should refer to. 
Taking the last example in Figure {\protect\ref{fig:main1}}, the \verb|addition| module would first compute the distribution over result list for `Albanian and Bulgarian citizens'. The \verb|subtraction| module can identify itself in the second step calculation and take the correct input to construct the new matrix $\textbf{C}_i'$. The expected distribution over new result list $RL'$ now represent the difference of `Greek citizens' and the previous result.

Instead of introducing specific modules for multi-number arithmetic such as `3-num-add', the structure of NMNs allows us to recursively execute basic operations several times in a compositional program. This design is in accord with the reasoning process of the CQA task, and natural for NMNs to perform complex computations.

%!TEX ROOT = ./coling2022.tex
\section{Experiments}
\paragraph{Dataset.}
% Unlike other CQA datasets, questions in DROP based on understanding of questions and paragraphs, which fits in well with the problem NMNs aim to solve.
We construct our own train/dev/test sets based on the DROP dataset \cite{dua2019drop}, which requires numerical reasoning skills.

\citet{Gupta2020NeuralMN} extracted a subset of questions from DROP that is supported by the model's reasoning capability. 
This subset contains approximately 20,000/500/2,000 QA pairs for train/dev/test. 
To train the \verb|add/sub| modules, we augment the NMNs' subset with more than 5,000 new questions from DROP. 
These questions were heuristically identified based on first n-grams and regular expressions (Appendix \ref{sec:appendix1}). 
Statistics of this newly constructed dataset can be found in Table \ref{table0}. 
Note that the \textsc{add-sub} questions include both 2-/and 3-number arithmetic and \textbf{all experiments in this paper are conducted on this new dataset}. 
Model performance is evaluated with the same F1 and EM (Exact Match) scores as \citet{Gupta2020NeuralMN}.

\begin{table}[h]
\centering
\resizebox{\columnwidth}{!}{
\begin{tabular}{lrrr}
\toprule
Question types    & train         & dev        & test       \\ \midrule
Full              & 25,165        & 623        & 2,547       \\ \midrule
\textsc{date-compare} (13.9\%)     & 3,505         & 91         & 333        \\
\textsc{date-difference} (12.2\%)  & 3,055         & 75         & 313        \\
\textsc{number-compare}  (12.1\%)  & 2,642         & 157        & 632        \\
\textsc{extract-number}  (12.8\%)  & 3,349         & 57         & 222        \\
\textsc{count} (17.3\%)            & 4,527         & 73         & 288        \\
\textsc{extract-argument} (13.1\%) & 3,467         & 51         & 208        \\
\textsc{\textbf{add-sub}} (18.6\%)  & \textbf{4,689}   & \textbf{124}  & \textbf{553}        \\ 
\quad{2-numbers} & 4,440         & 106         & 505        \\  
\quad{3-numbers}  & 259           & 24         & 66        \\  \bottomrule
\end{tabular}
}
\caption{Question types distribution on the expanded DROP subset used in the follow experiments.}
\label{table0}
\end{table}

\paragraph{Result.}
\noindent In Table \ref{table1}, we list the overall performance of the original NMNs, NumNet and our proposed method NMNs$_\pm$.

\begin{table}[htb]
\centering\small
\begin{tabular}{lrr}
\toprule
Method                                 & F1   & EM   \\ \midrule
original NMNs \cite{Gupta2020NeuralMN} & 57.5 & 54.9 \\ 
NumNet \cite{2019NumNet}               & 72.3 & 69.4 \\ \midrule
NMNs$_\pm$ (ours)                      & \textbf{75.2} & \textbf{72.6} \\ 
 \quad w/o add/sub                     & 61.4 & 58.1 \\ 
 \quad w/o qi                           & 74.3 & 71.7 \\
\bottomrule
\end{tabular}
\caption{Performance comparison between different models on \textbf{our test set}. Constrained by the page limit, case study and analysis are in Appendix \ref{sec:appendix3}.}
\label{table1}
\end{table}

In Table~\ref{table1}, row ``w/o add-sub'' is the model variant with question attention only, and row ``w/o qi'' only has the \verb|add/sub| modules only. 
Compared to the original NMNs, two proposed methods both improve model performance and the \verb|add/sub| modules contributes more.
Our full NMNs$_\pm$ model, with both components added, achieves 75.2\% F1 and 72.6\% EM scores, obtaining significant deltas of 17.7\% absolute points compared to the original NMNs for both F1 and EM. 
Additionally, NMNs$_\pm$ outperforms NumNet by 2.9\% and 3.2\% absoule points in F1 and EM.
% Note that the original NMNs achieve the 77.4\% F1 and 74.0\% EM on the original subset \cite{Gupta2020NeuralMN}, which does not include addition and subtraction questions. 
%It can be observed from the first row in Table~\ref{table1} that the orginal NMNs' performance on the new dataset declines substantially (57.5 F1 and 54.9 EM), because the original model did not train to solve addition and subtraction questions. 

% We study the contribution of each component in an ablation analysis. 

It can be unfair since the original NMNs will perform poorly on the newly added \textsc{add-sub} questions. 
Therefore, we list the model performance on different question types in Table \ref{table2}. 
Our model achieves higher scores across almost all question types comparing to the original NMNs, attesting to the effectiveness of our proposed techniques. 
And it turns out that adding ADD-SUB question types and more training data does not improve the results of the original DROP split.
This might due to the performance degradation of the programmer after adding these new ADD-SUB programs.
When comparing to NumNet, though our model fail on 2-number \textsc{add-sub} questions, we achieve 5.4\% F1 improvement on 3-number \textsc{add-sub} questions, thus results in a comparable performance.
Note that the 2-number data is nearly \textbf{18 times} the 3-number data, which shows our model or NMNs relies less on large scale datasets.

\begin{table}[hbt!]
\centering
% \scalebox{0.60}{
\resizebox{\columnwidth}{!}{
\begin{tabular}{lrrr}
\toprule
Question type       & NMNs & NMNs$_\pm$ & NumNet \\ \midrule
\textsc{date-compare}       & 79.2          & \textbf{84.9}   & 72.0        \\
\textsc{date-difference}   & 69.0          & 73.3 & \textbf{74.1}           \\
\textsc{number-compare}     & 89.6          & \textbf{90.3}  & 89.9          \\
\textsc{extract-number}      & 86.4         & \textbf{89.1}   & 85.6        \\
\textsc{count}               & 54.2          & \textbf{60.2}  & 52.4         \\
\textsc{extract-argument}     & 73.4           & \textbf{75.3}   & 66.1        \\
\textsc{add-sub}    & 0.7          & 66.4     & \textbf{67.6}     \\
\quad {2-numbers}    & 0.8          & 67.9 & \textbf{71.5}          \\
\quad {3-numbers}   & 0.3          & \textbf{41.2}   & 35.8         \\\bottomrule
\end{tabular}
}
\caption{F1 comparison on different question types.}
\label{table2}
\end{table}

% Moreover, NMNs$_\pm$ dramatically improve arithmetic performance from 0.7\% to 66.4\%. 
Additional ablation studies for the \verb|add/sub| modules (\ref{sec:appendix2}) and a qualitative analysis (\ref{sec:appendix4}) can be found in the appendix.

%!TEX ROOT = ./coling2022.tex
\section{Conclusion}
In this work, we extend NMNs' numerical reasoning capability to 2-/and 3-number addition and subtraction, and incorporate the influence of question information to the interpreter on number related questions. 
Experimental results show that our methods significantly enhance NMNs' numerical reasoning ability, with an increase of 17.7\% absolute F1 points on a newly constructed DROP subset that includes arithmetic questions.
Moreover, our approach also outperforms NumNet, a SOTA numerical reasoning model, by 2.9\% F1 points.

%!TEX ROOT = ./coling2022.tex
\section*{Acknowledgements}
This work is partially funded by the DARPA CCU program (HR001121S0024).

\bibliography{custom,anthology}
\bibliographystyle{acl_natbib}

%!TEX ROOT = ./coling2022.tex
\clearpage
\newpage

\appendix
\section{Appendix}

\subsection{Hyper-parameter setting for compare-date modules}
\label{sec:appendix0}
\noindent As mentioned above, we use a hyper-parameter $\alpha$ to represent question's and paragraph's weights for the combined context representation. We determine the final coefficient through a series of control parameter comparison experiments: use the same data to train and validate the model with different $\alpha$. The model achieves the best performance (84.9 F1) for \textsc{date-compare} questions when $\alpha$ was set to 0.4 (40\% for paragraph attention and 60\% for question attention), which increase 5.7 absolute points compared to the original NMNs model. The experiment verifies the importance of question information in the numerical reasoning process.

\subsection{Data extraction}
\label{sec:appendix1}
In this research, we expand the DROP subset for original NMNs to cover addition and subtraction questions. Subtraction questions can be easily targeted by their first n-gram, such as `how many more', `how many yards difference'. For three number subtraction, we need to further specified the format by regular expression, such as `how many more \textsc{event-a} and \textsc{event-b} than \textsc{event-c}?' or `how many more \textsc{event-a} compared to \textsc{event-b} and \textsc{event-c}?'. For addition, it is hard to identify how many numbers should participate in the calculation from some of the questions (e.g. `how many total yards did Roethlisberger get in the game?'). Therefore, we use regular expression to distinguish two or three numbers addition and follow the patterns such as `how many total...', `how many ... combined'.

\subsection{Addition and subtraction modules training}
\label{sec:appendix2}
\noindent To discuss the contribution of individual \verb|addition| and \verb|subtraction| module for NMNs, we conduct an ablation experiment by training and testing the model on different datasets as shown in Table \ref{table3}. The five rows represent the model trained on various datasets: addition questions only, subtraction questions only, addition and the original NMNs subset, subtraction and original NMNs subset and our full subset. The columns indicate the model performance results when they test on addition/subtraction questions only and the full DROP subset. As can be seen from the result, the model with subtraction ability only perform greater than with addition ability only.

\begin{table}[hbt!]
\centering
\begin{tabular}{lllll}
\hline
\multirow{2}{*}{Datasets}          & \multicolumn{2}{l}{addsub dataset} & \multicolumn{2}{l}{full dataset} \\
                                   & F1                 & EM                 & F1                     & EM                    \\ \hline
add                 & 41.2               & 41.2               & 46.0                   & 43.8                  \\
sub              & 45.7               & 45.7               & 51.3                   & 49.2                  \\
add+origin             & 51.5               & 51.5               & 69.2                   & 64.2                  \\
sub+origin          & 55.1               & 55.1               & 72.6                   & 69.8                  \\
add+sub+origin & 66.4               & 66.4               & 74.3                   & 71.7                  \\ \hline
\end{tabular}
\caption{Ablation experiment result for addition and subtraction modules}
\label{table3}
\end{table}

\subsection{Qualitative analysis}
\label{sec:appendix4}
Figure \ref{fig:appendix1} shows some incorrect prediction cases from the original NMNs and the answer from our improved model NMNs$_\pm$. From the examples, we can clearly identify how the proposed techniques improve the numerical reasoning process:

\begin{figure*}[hbt!]
  \centering
  \includegraphics[width=1.0 \textwidth]{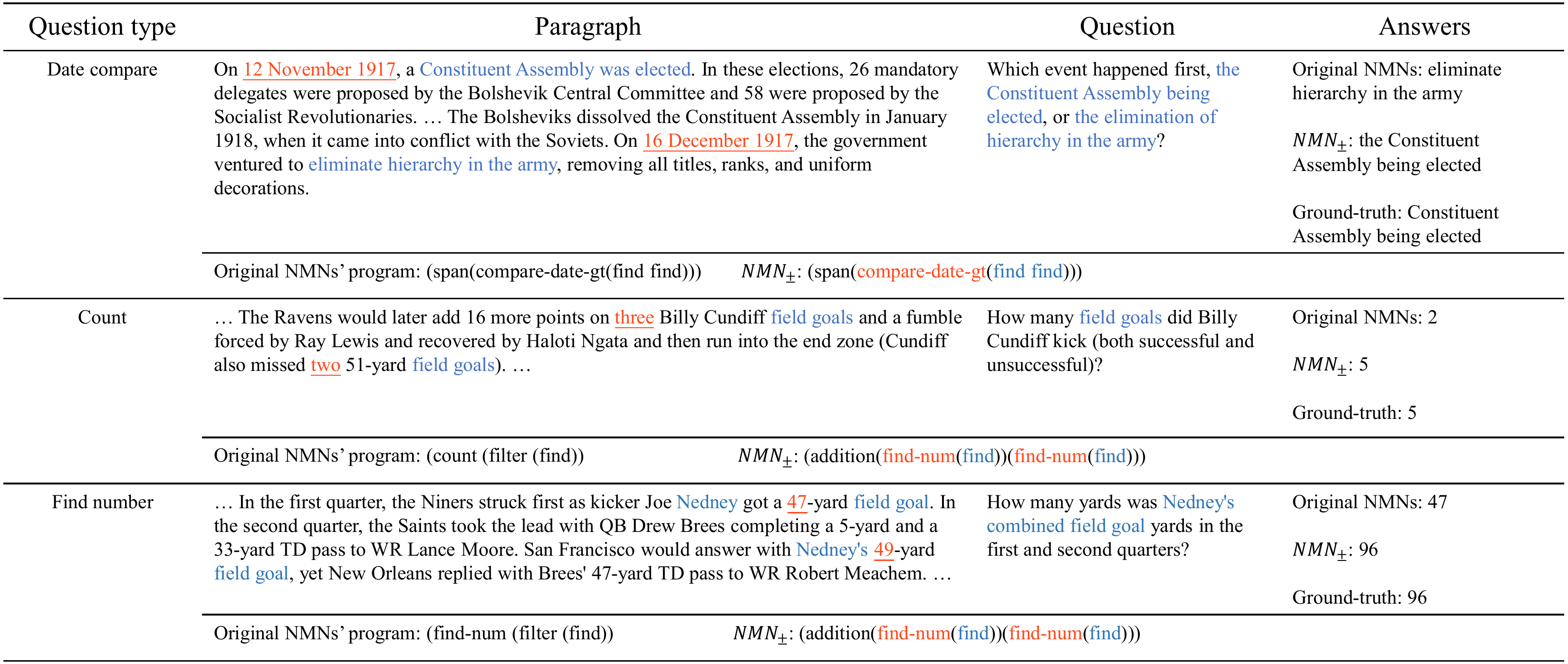}
  \caption{Qualitative analysis. The highlighted spans are corresponding to the modules in the program for each question.}
  \label{fig:appendix1}
\end{figure*}

\begin{itemize}
    \item In the first example, the original NMNs match wrong tokens `dissolved the Constituent Assembly' given the question `Which event happened first, the Constituent Assembly being elected, or the elimination of hierarchy in the army?', thus located a wrong date `January 1918'. After enhancing the interpreter's awareness of the question, NMNs$_\pm$ can precisely target the spans `a Constituent Assembly was elected' in the paragraph and further provide the correct prediction.
    \item The following two examples are wrongly answered by the original NMNs because of incorrect program predictions. The second question was initially categorized into a \textsc{count} question which called the \verb|count| module to calculate the number of attended paragraph spans. The same situation occurs in the third question, because the original NMNs lack the modules that can correctly expresses the reasoning behind the question. The prediction results prove that our NMNs$_\pm$ model handle simple arithmetic operations such as addition and subtraction which meets the task requirement.
\end{itemize}

\subsection{Prediction analysis}
\label{sec:appendix3}
The wrong prediction cases study for the original NMNs over DROP is the main motivation of our proposed methods. We conclude the error factors of five numerical question types in detail: \textsc{date-compare}, \textsc{count}, \textsc{date-difference}, \textsc{number-compare} and \textsc{extract-number}.

\begin{figure*}[hbt!]
  \centering
  \includegraphics[width=1.0 \textwidth]{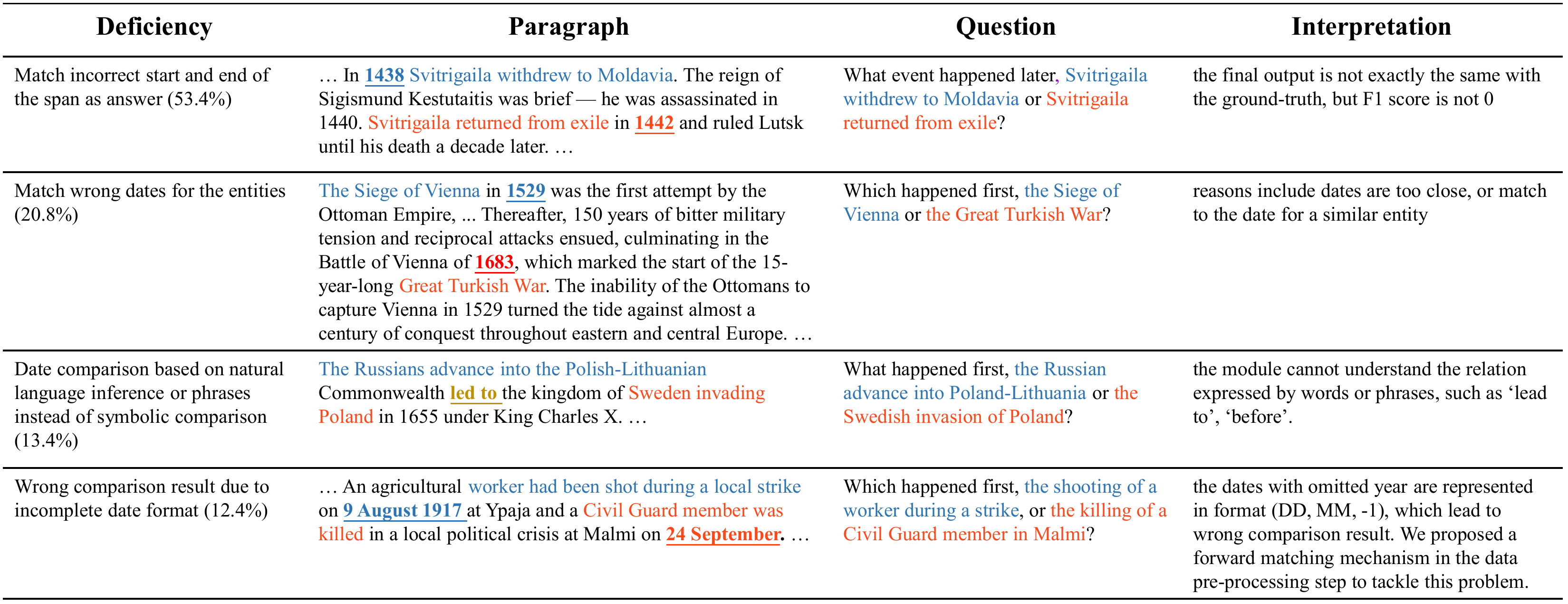}
  \caption{Root causes for the wrong prediction in \textsc{date-compare} questions. The related events mentioned in the question are highlighted in blue and red, and their relevant dates are in the same color with underline.}
  \label{fig:appendix2}
\end{figure*}

\begin{figure*}[hbt!]
  \centering
  \includegraphics[width=1.0 \textwidth]{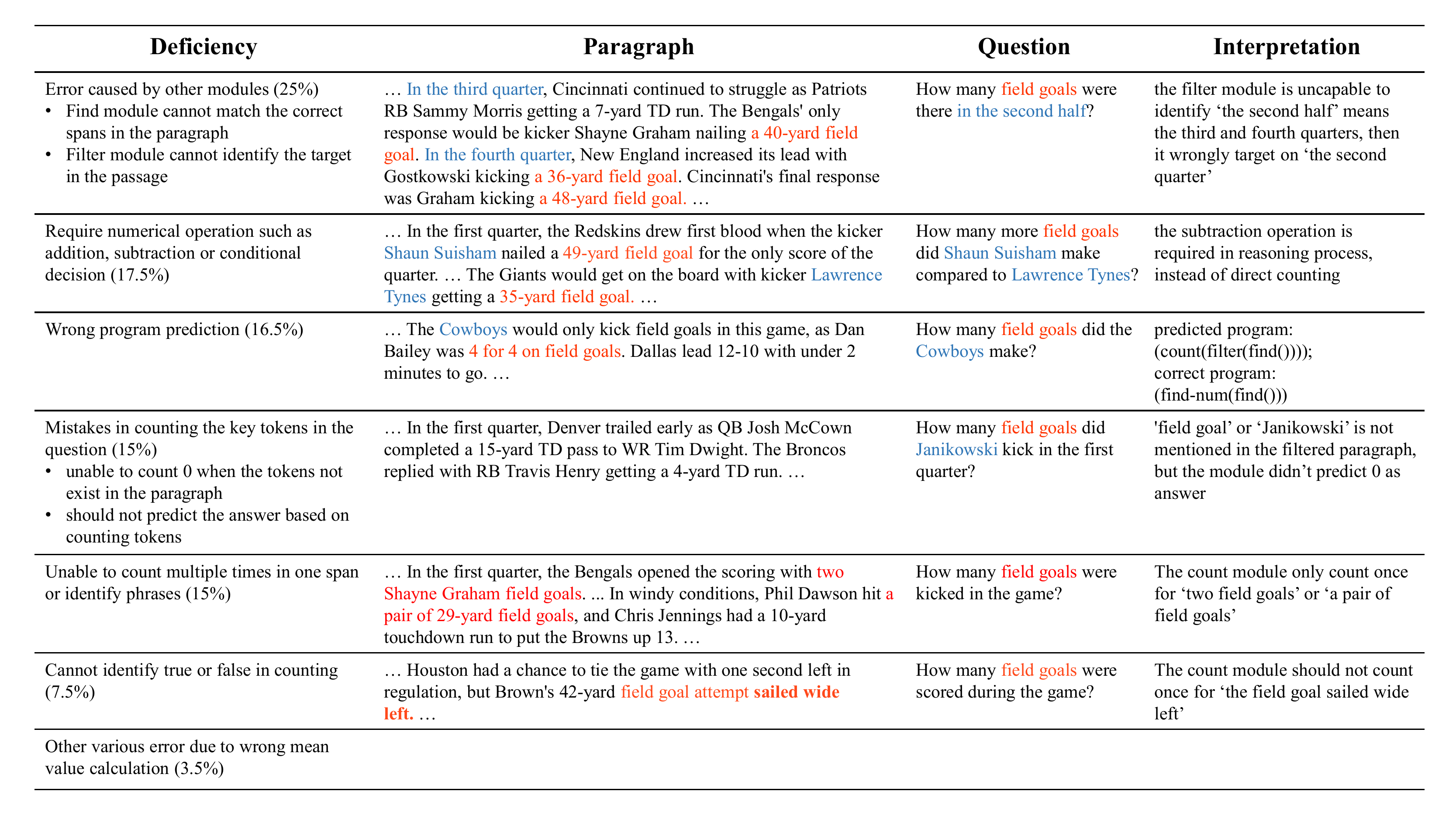}
  \caption{Root causes for the wrong prediction in \textsc{count} questions. The inputs to the find module and their targets in the paragraph are highlighted in red. The blue spans are related to the filter module.}
  \label{fig:appendix3}
\end{figure*}

\begin{figure*}[hbt!]
  \centering
  \includegraphics[width=1.0 \textwidth]{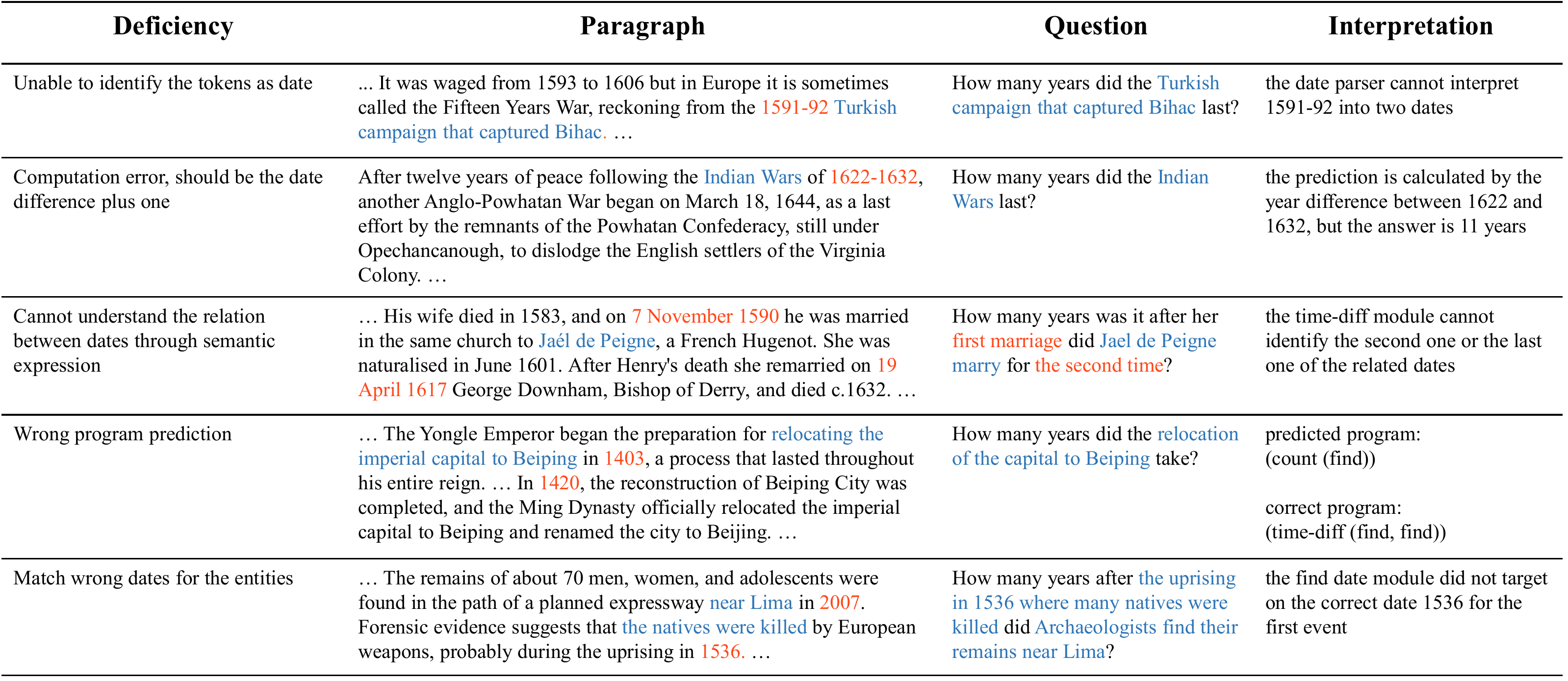}
  \caption{Root causes for the wrong prediction in date-difference questions. The related events are highlighted in blue, which is the input of the find module. The dates grounding correctly predicted in the compare-date modules are highlighted in red color. The answer predicted by NMNs should be the difference of these two dates.}
  \label{fig:appendix4}
\end{figure*}

\begin{figure*}
  \centering
  \includegraphics[width=1.0 \textwidth]{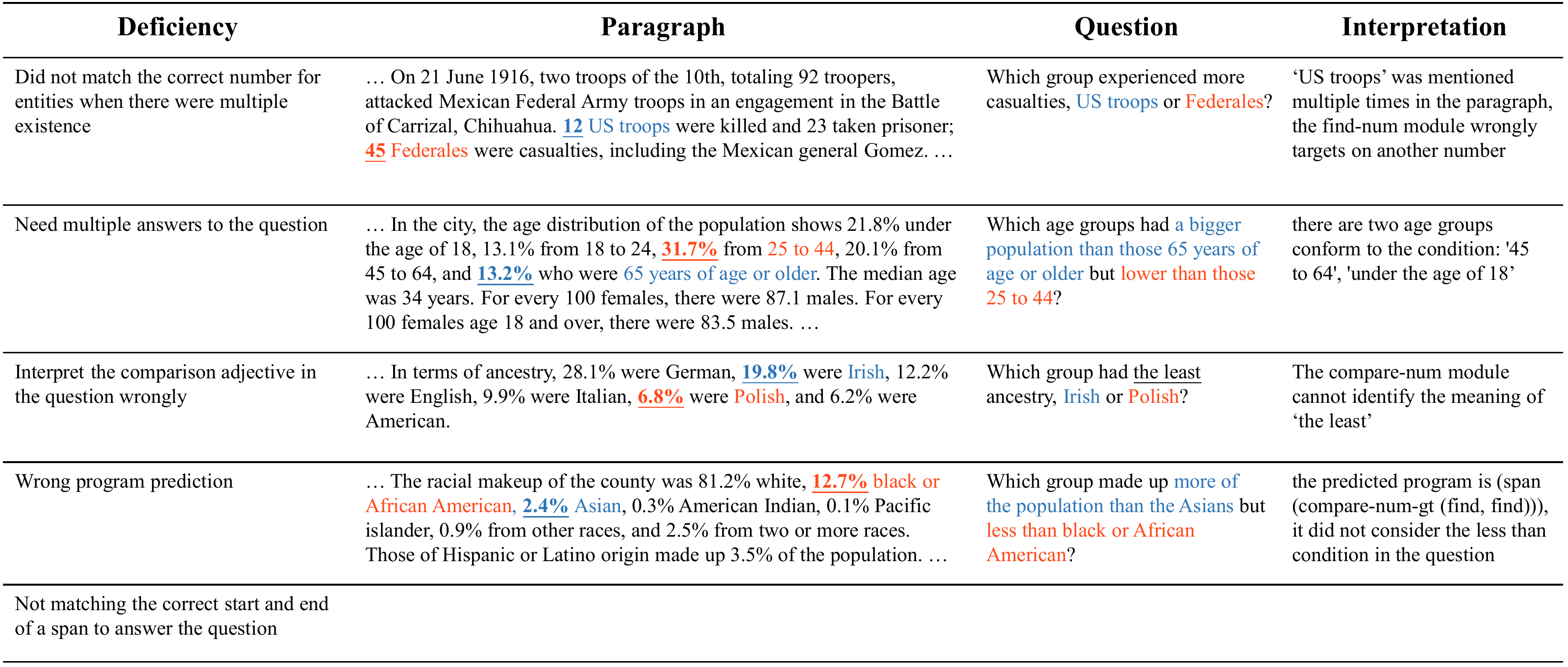}
  \caption{Root causes for the wrong prediction in number-compare questions. Similar to figure 1, the input of the find module is highlighted in blue and red, and their related numbers are underlined. The paragraph span predicted as the answer is the one associated to a smaller/larger-valued number according to the questions asking.}
  \label{fig:appendix5}
\end{figure*}

\begin{figure*}
  \centering
  \includegraphics[width=1.0 \textwidth]{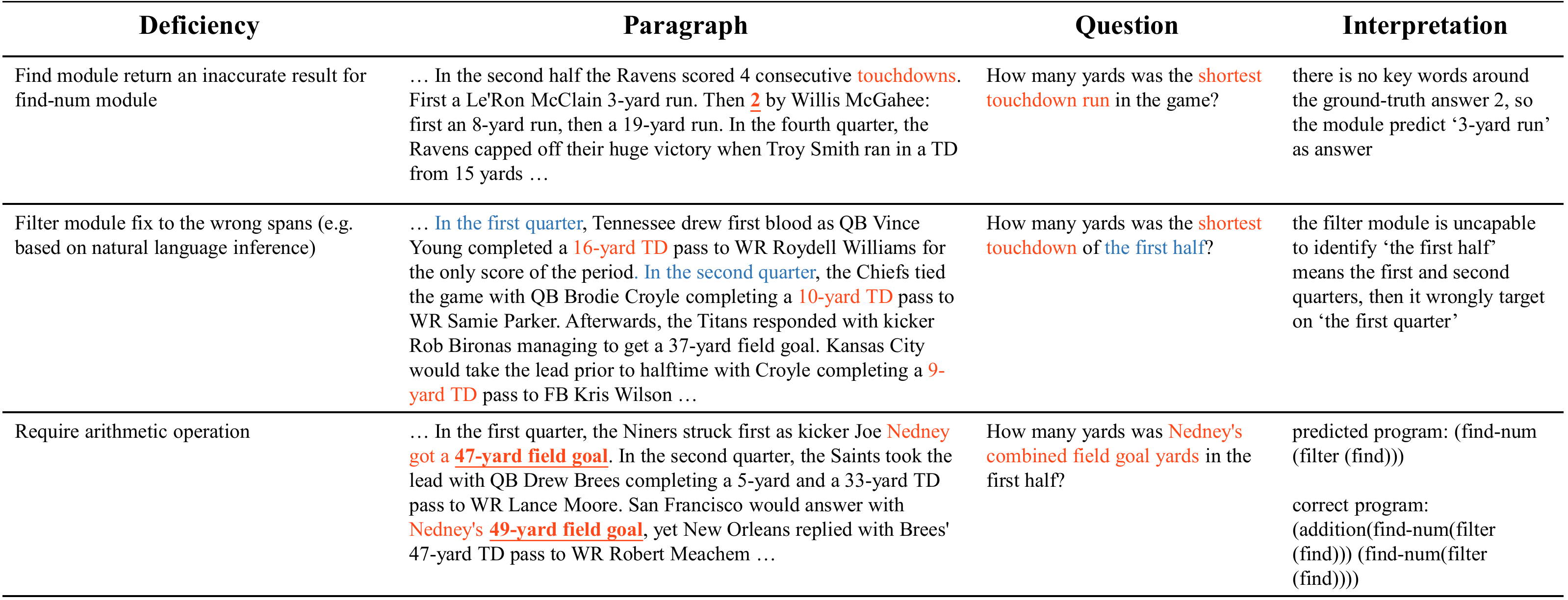}
  \caption{Root causes for the wrong prediction in extract-number questions. The inputs to the find module and their targets in the paragraph are highlighted in red. The blue spans are related to the filter module. The find-num module finally extracts the number associated with this paragraph attention as the answer.}
  \label{fig:appendix6}
\end{figure*}

\end{document}